\documentclass[10pt,twocolumn,letterpaper]{article}

\usepackage{cvpr}              
\definecolor{cvprblue}{rgb}{0.21,0.49,0.74}
\usepackage[pagebackref,breaklinks,colorlinks,allcolors=cvprblue]{hyperref}
\usepackage[utf8]{inputenc} 
\usepackage[T1]{fontenc}    
\usepackage{hyperref}       
\usepackage{url}            
\usepackage{booktabs}       
\usepackage{amsfonts}       
\usepackage{nicefrac}       
\usepackage{microtype}      
\usepackage{xcolor}         
\usepackage{graphicx}
\usepackage{subcaption}
\usepackage{tabularx}
\usepackage{pdflscape}
\usepackage{pifont}
\usepackage{comment}
\usepackage{amsmath}
\usepackage{enumitem}
\usepackage{xcolor}
\usepackage[most]{tcolorbox}
\definecolor{myPurple}{RGB}{128,0,128}
\definecolor{clsBus}{RGB}{25,70,190}
\definecolor{clsBike}{RGB}{205,95,0}
\definecolor{clsCar}{RGB}{20,135,55}
\definecolor{clsPed}{RGB}{200,20,40}
\usepackage{pgfplots}
\pgfplotsset{compat=1.18}

\hypersetup{
    colorlinks=true,
    linkcolor=blue,
    urlcolor=blue,
    citecolor=green,
}

\def\paperID{*****} 
\def\confName{CVPR}
\def\confYear{2026}

\title{Towards Continual Expansion of Data Coverage: \\ Automatic Text-guided Edge-case Synthesis}

\author{Kyeongryeol Go\\
Superb AI\\
Seoul, South Korea\\
{\tt\small krgo@superb-ai.com}
}

\begin{document}
\maketitle
\begin{abstract}
    The performance of deep neural networks is strongly influenced by the quality of their training data. However, mitigating dataset bias by manually curating challenging edge cases remains a major bottleneck. To address this, we propose an automated pipeline for text-guided edge-case synthesis. Our approach employs a Large Language Model, fine-tuned via preference learning, to rephrase image captions into diverse textual prompts that steer a Text-to-Image model toward generating difficult visual scenarios. Evaluated on the FishEye8K object detection benchmark, our method achieves superior robustness, surpassing both naive augmentation and manually engineered prompts. This work establishes a scalable framework that shifts data curation from manual effort to automated, targeted synthesis, offering a promising direction for developing more reliable and continuously improving AI systems. Code is available in our \href{https://github.com/gokyeongryeol/ATES}{\textcolor{magenta}{GitHub repository}}.
\end{abstract}    
\section{Introduction}
\label{sec:intro}

\begin{figure*}[ht]
    \centering
    \begin{tabular}{cccc}
        \makebox[0.19\linewidth]{\textbf{real}} &
        \makebox[0.19\linewidth]{\textbf{naive}} &
        \makebox[0.19\linewidth]{\textbf{manual}} &
        \makebox[0.19\linewidth]{\textbf{automatic (ours)}} \\
        \includegraphics[width=0.19\linewidth]{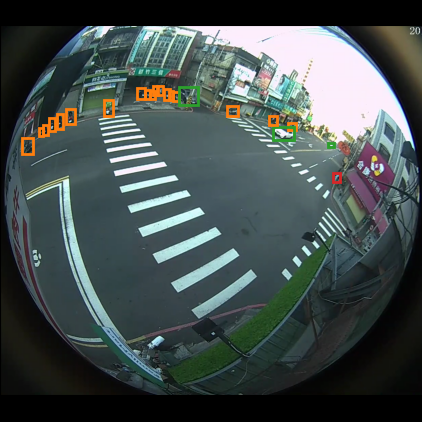} &
        \includegraphics[width=0.19\linewidth]{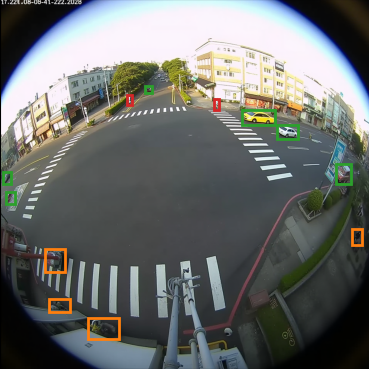} &
        \includegraphics[width=0.19\linewidth]{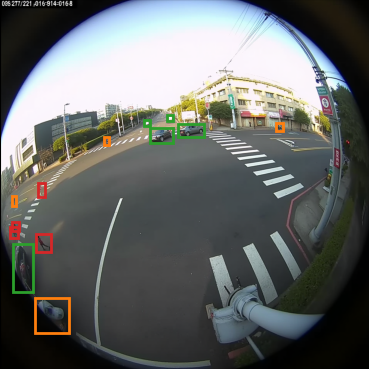} &
        \includegraphics[width=0.19\linewidth]{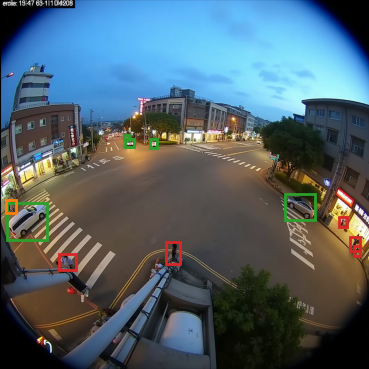} \\
        \includegraphics[width=0.19\linewidth]{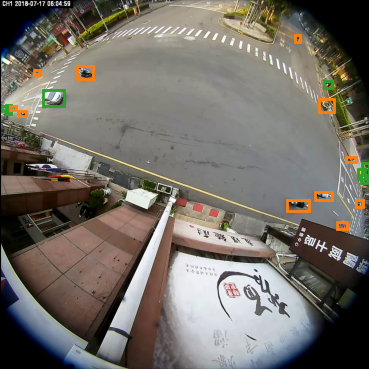} &
        \includegraphics[width=0.19\linewidth]{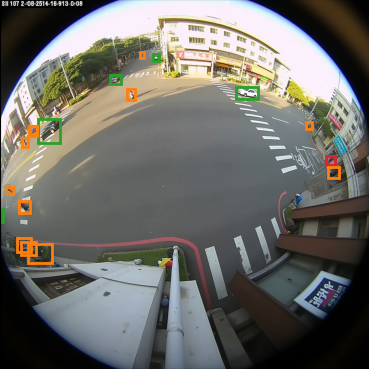} &
        \includegraphics[width=0.19\linewidth]{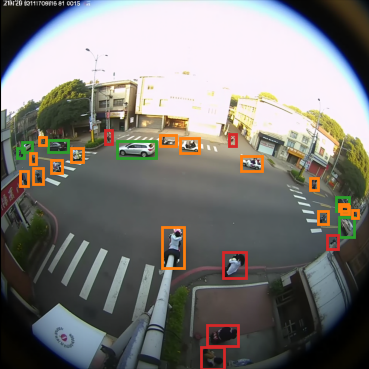} &
        \includegraphics[width=0.19\linewidth]{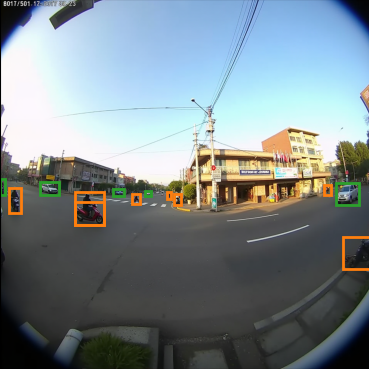} \\
        \includegraphics[width=0.19\linewidth]{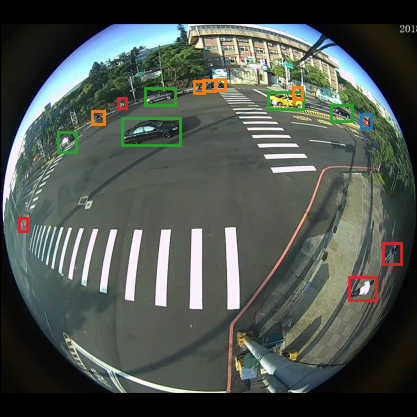} &
        \includegraphics[width=0.19\linewidth]{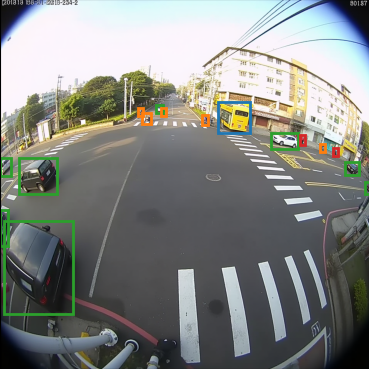} &
        \includegraphics[width=0.19\linewidth]{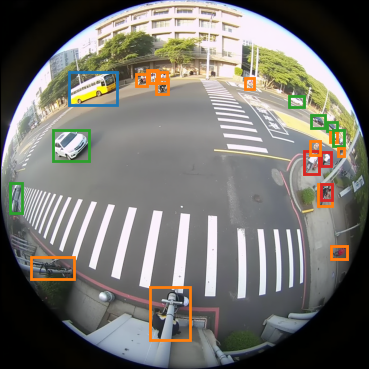} &
        \includegraphics[width=0.19\linewidth]{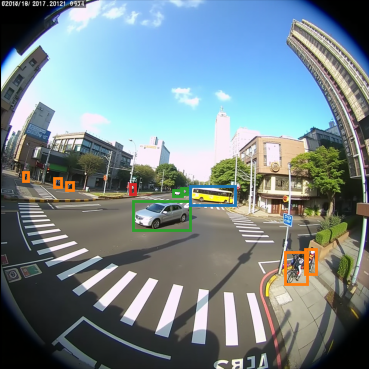} \\
    \end{tabular}
    \caption{Comparison of synthetic data generation strategies. Each row displays examples generated from the same real image base caption using three different approaches: naive, manual, and automatic (ours). This row-wise arrangement isolates the effect of the generation strategy on the resulting visual output and its associated labels. Note that pseudo annotations are marked with different colors indicating different classes (\textcolor{clsBus}{Bus}, \textcolor{clsBike}{Bike}, \textcolor{clsCar}{Car}, \textcolor{clsPed}{Pedestrian}, \textcolor{myPurple}{Truck}). Best viewed in color.}
    \label{fig: syn_examples}
\end{figure*}

The performance of deep neural networks is fundamentally dependent on the quality and diversity of training data. However, real-world datasets are often fraught with noisy labels and inherent biases, which can significantly impair model generalization and robustness \citep{hort2024bias}. To mitigate these issues, prior research has largely pursued two complementary directions: model-centric methods \citep{goldberger2017training, zhang2018generalized, liu2020early, nam2020learning, li2020dividemix}, such as robust optimization and regularization techniques, and data-centric methods \citep{park2020swapping, lee2021learning, ahn2022mitigating}, including data augmentation and re-sampling strategies.

These methods help mitigate deficiencies in existing datasets. However, they remain largely remedial in nature. A growing trend instead focuses on synthesizing new data to capture model weaknesses and underrepresented edge-case scenarios. Traditionally, identifying such edge-cases has been a labor-intensive process. It typically requires domain experts to manually analyze model failure modes, scrutinizing confusion matrices, class-wise performance metrics, and specific false positives and negatives, to guide subsequent data collection \citep{kim2025edge}. This manual intervention renders the process not only time-consuming and expertise-dependent but also difficult to automate, scale, and reproduce.

To overcome these limitations, we introduce a novel, automated pipeline for targeted edge-case synthesis. Our pipeline employs a Large Language Model (LLM) fine-tuned with preference learning. This LLM is trained to systematically rephrase image captions from an existing dataset into textual descriptions that characterize challenging edge-case scenarios. These newly generated captions are then fed as prompts to a pre-trained text-to-image (T2I) model, yielding a curated set of diverse and challenging training examples designed to bolster model robustness.

A key distinction of our work lies in its approach to identifying edge-cases at the caption level, rather than through the analysis of image or instance embeddings as in prior work \citep{du2023dream, zhang2023expanding, hayden2025generative}. This design choice is motivated by two key insights. First, we find that simple semantic modifications to captions can induce substantial and meaningful diversity in the generated images, affecting not only object attributes and arrangements but also the broader scene context and background. Second, by building upon the standard T2I generation framework, our pipeline is poised to readily incorporate future advancements in high-fidelity generative models and parameter-efficient fine-tuning techniques.

By automating the discovery and synthesis of edge-cases through generative captioning, our work provides a scalable and efficient solution for improving model performance on challenging, real-world scenarios. Our main contributions are summarized as follows:
\begin{itemize}
\item We propose a novel pipeline that automates the generation of targeted edge-case data by leveraging a preference-tuned LLM to create challenging prompts for a T2I model.
\item We demonstrate that training with our synthesized data enhances model robustness and generalization on critical edge-case scenarios compared to existing methods.
\end{itemize}

\section{Related work}
\label{sec: related}

\begin{figure*}[ht]
    \centering
    \includegraphics[width=\textwidth]{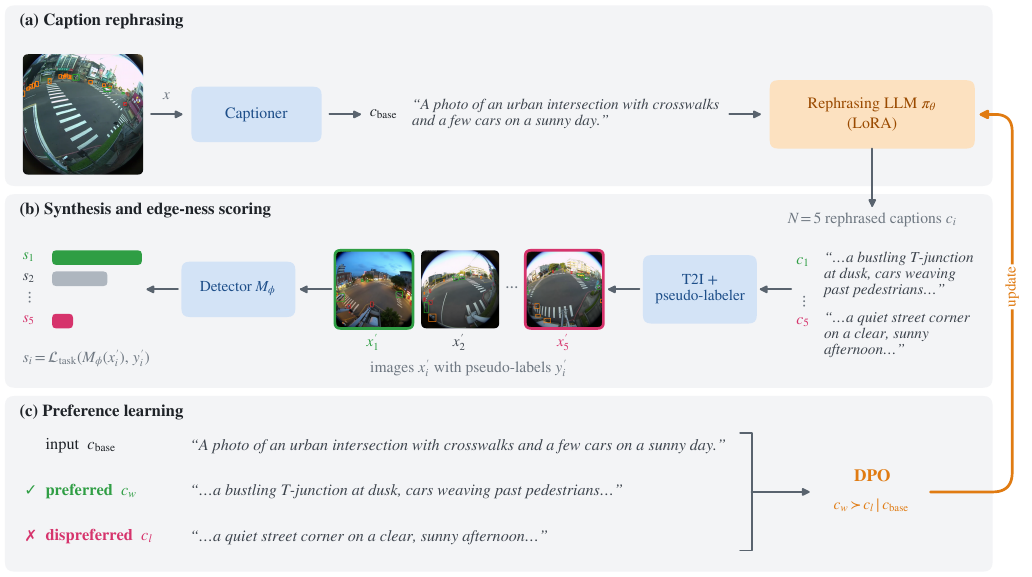}
    \caption{Training pipeline of the rephrasing LLM to understand and generate edge-case-aware captions. This cyclical process teaches the LLM to generate prompts that are more likely to produce challenging images for the discriminative model. Best viewed in color.}
    \label{fig: pipeline_train}
\end{figure*}

\paragraph{Synthetic data from generative models.}
Recent advancements in generative models, particularly diffusion-based approaches, have led to a remarkable improvement in the quality of synthetic data generation. This progress has spurred active research into leveraging these models for data augmentation, with generated data enhancing the performance of various visual recognition tasks such as classification \citep{azizi2023synthetic}, detection \citep{chen2023geodiffusion}, and segmentation \citep{nguyen2023dataset}.

While generative models offer a promising solution to imperfect data, their application is still in challenges. A primary concern is their tendency to generate data centered around the main modes of the training distribution \citep{yamaguchi2023limitation}, a phenomenon often referred to as mode-collapse. This issue is particularly acute when dealing with real-world datasets that inherently exhibit a long-tail distribution, where over-reliance on typical samples can be detrimental to the model's ability to handle infrequent but critical cases \citep{zhang2023deep}. In essence, beyond a certain threshold, the utility of synthetic data diminishes as it fails to capture the diversity of rare, edge-case scenarios \citep{fan2024scaling}. Therefore, to effectively enhance the robustness and accuracy of discriminative models, a more strategic approach is required---one that progressively expands the data distribution coverage, with a specific focus on generating synthetic samples in regions where the model exhibits high uncertainty.

\paragraph{Automated synthesis of challenging data.}
A significant body of research has focused on automatically generating challenging or out-of-distribution (OOD) data to improve model generalization. Early approaches, such as DREAM-OOD \citep{du2023dream}, learn a compact, text-conditioned latent space from in-distribution (ID) data. Outliers are then synthesized by sampling new embeddings from the low-likelihood regions of this learned manifold and then decoding them into photorealistic images. This method targets the boundaries of the known data distribution to produce novel samples.

Subsequent research shifted from generating generic outliers to creating more informative samples specifically tailored for model training. The Guided Imagination Framework (GIF) \citep{zhang2023expanding}, for instance, expands small-scale datasets by performing optimized perturbations in the latent space of a pre-trained generative model. The optimization is guided by dual objectives: maximizing the information content of samples (e.g., by increasing classification entropy) while preserving their original class identity and promoting sample diversity. More recently, Longtail-Guided Diffusion (LTG) \citep{hayden2025generative} proposed to condition the generation process directly on a target model's blind spots. LTG introduces a differentiable, model-based longtail signal via an Epistemic Head, which guides a latent diffusion model during inference to synthesize instances where the model is uncertain.

Despite their ingenuity, these methods often rely on navigating complex latent spaces and can be constrained by restrictive structural assumptions. Their performance is frequently sensitive to hyperparameter tuning, and the interpretability of the generated data can be limited, posing challenges for scalability and practical deployment.

\paragraph{Data-centric and prompt-based generation.}
In parallel with automated, model-centric techniques, another line of work has demonstrated the efficacy of a more hands-on, data-centric pipeline, particularly in specialized domains. A notable example is found in fisheye object detection \citep{kim2025edge}, where researchers conducted a detailed manual error analysis to identify critical edge-cases, such as confusing class pairs or objects distorted at the image periphery. These qualitative insights were then translated into systematically crafted text prompts, which guided a fine-tuned T2I model to synthesize images replicating these specific failure modes.

While this approach has proven highly effective for targeted problem-solving, its core limitation is its dependence on domain expertise and manual prompt engineering. This reliance makes the process difficult to scale and adapt to new domains or models without significant human intervention.

Our work aim to combine the scalability and automation of model-conditioned guidance with the semantic precision of targeted, prompt-based generation. Instead of navigating latent embeddings or requiring manual prompt crafting, we propose an automated method to discover and formulate edge-cases at the caption level, thereby creating a scalable yet highly targeted data synthesis pipeline.
\section{Methodology}
\label{sec:method}

The overall pipeline is illustrated in Figure~\ref{fig: pipeline_train}. We aim to automatically synthesize high-value, edge-case data, thereby progressively expanding the coverage and robustness of a training dataset. The core of our methodology is a self-improving feedback loop wherein a LLM is fine-tuned to rephrase image captions, guiding a generative model toward producing images that are challenging for a given discriminative model. This process leverages preference learning to align the LLM's rephrasing strategy with a quantitative measure of edge-ness. 

\subsection{Initial data preparation}
The process commences with an image $x$ from the original training dataset $D$. First, a pre-trained image captioning model, $Cap(\cdot)$, generates a concise descriptive caption, $c_{base} = Cap(x)$. This base caption serves as a factual, high-level summary of the image content.

Next, to explore the semantic space around this description, $c_{base}$ is fed into a rephrasing LLM, denoted as $\pi_{\theta}$, where $\theta$ are its parameters. The LLM is prompted to produce a set of $N$ diverse semantic variations, $C=\{c_1, c_2, ..., c_N\}$, where each $c_i \sim \pi_{\theta}(\cdot \vert c_{base})$. These rephrased captions are designed to alter contextual details, object attributes, or narrative perspectives of the original scene, forming the foundation for generating a diverse range of visual outputs. While increasing $N$ enhances the diversity of the rephrased captions, it also incurs additional computational cost, as each caption must be synthesized and pseudo-labeled in the following steps. For practicality, we fix $N=5$.

\subsection{Quantifying edge-ness for preference labeling}
Each of the $N$ rephrased captions $c_i \in C$ is passed to a pre-trained T2I model, $G(\cdot)$, to synthesize a corresponding image, $x_i' = G(c_i)$. Since these synthetic images lack ground-truth annotations, we employ a high-performance, pre-trained model as a pseudo-labeler, $PL(\cdot)$, to generate pseudo-annotations $y_i' = PL(x_i')$.

To evaluate which rephrased captions are most effective at generating challenging scenarios, we introduce the concept of edge-ness. We define edge-ness as a quantifiable measure of the difficulty a synthetic image poses to a discriminative model, $M_{\phi}$, which is pre-trained on the original training dataset $D$. Specifically, for each generated pair $(x_i', y_i')$, we compute the task-specific training loss: 

\begin{equation}
    s_i = \mathcal{L}_{task} \left ( M_{\phi}(x_i'), y_i' \right )
\end{equation}

A higher value of $s_i$ is interpreted as an indicator of edge-ness, as it signifies a greater discrepancy between the model's predictions and the pseudo annotations.

\subsection{Edge-case-aware rephrasing}
To align the LLM with the objective of generating challenging scenarios, we construct a preference dataset $D_{pref}$ consisting of pairs $(c_w, c_l)$, where $c_w$ denotes the caption that induced the maximum task loss (preferred) and $c_l$ denotes the caption that induced the minimum task loss (un-preferred) among the N variations generated for a given base caption. This dataset is used to fine-tune the rephrasing LLM $\pi_\theta$, effectively aligning its behavior toward generating captions that are more likely to produce high edge-ness images. For this preference-based fine-tuning, we adopt Direct Preference Optimization (DPO) \citep{rafailov2023direct}, which optimizes the model to increase the relative log-probability of $c_w$ over $c_l$ using the following objective:

\begin{equation}
\begin{aligned}
\mathcal{L}_{DPO}
&=
\log \sigma
\left(
\beta \big(
r_\theta(c_w \mid c_{base})
-
r_\theta(c_l \mid c_{base})
\big)
\right), \\
& \text{where} \quad
r_\theta(c \mid c_{base})
=
\log
\frac{\pi_{\theta}(c \mid c_{base})}
{\pi_{ref}(c \mid c_{base})}.
\end{aligned}
\end{equation}

Here, the reference model $\pi_{ref}$ is a frozen copy of the initial language model, $\sigma$ denotes the sigmoid function, and $\beta$ controls the strength of the preference signal.

Once the LLM is aligned, the newly fine-tuned model $\pi_{\theta'}$ is used to generate a set of edge-case-aware captions. These captions are used to synthesize new images, which, along with their pseudo annotations, are used to augment the original training set: $D_{aug} = D \cup \{(x_{new}', y_{new}')\}$. Please refer to Appendix~\ref{sec: aug_pipeline} for the data augmentation pipeline.

Notably, this entire pipeline can be executed iteratively. By retraining the discriminative model $M_{\phi}$ on $D_{aug}$ and using the updated model as the next-stage edge-ness scorer, the system can progressively discover and synthesize increasingly complex edge-cases, continually expanding the data coverage.
\section{Experiment}
\label{sec:experiments}

\begin{table*}[t]
\centering
\caption{YOLOv11-small performance under iterative data augmentation strategies, comparing naive, manual, and automatic pipelines, with evaluation using mAP and mAP w/o TP.}
\begin{tabular}{lccc}
\toprule
Train dataset & Count & mAP & mAP w/o TP \\
\midrule
train-\textit{D} & 3,187 & 0.335 $\pm$ 0.005 & - \\
train-\textit{D} + naive v0 & 19,122 & 0.376 $\pm$ 0.004 & - \\
\hline
train-\textit{D} + naive v0 + naive v1 & 35,057 & 0.378 $\pm$ 0.002 & 0.361 $\pm$ 0.002 \\
train-\textit{D} + naive v0 + manual v1 & 35,057 & 0.374 $\pm$ 0.003 & 0.357 $\pm$ 0.003 \\
train-\textit{D} + naive v0 + automatic v1 & 35,057 & \underline{0.381 $\pm$ 0.003} & \underline{0.363 $\pm$ 0.003} \\
train-\textit{D} + naive v0 + naive v1 + naive v2 & 50,992 & 0.380 $\pm$ 0.002 & 0.362 $\pm$ 0.003 \\
train-\textit{D} + naive v0 + manual v1 + manual v2 & 50,992 & 0.380 $\pm$ 0.001 & 0.362 $\pm$ 0.001 \\
train-\textit{D} + naive v0 + automatic v1 + automatic v2 & 50,992 & \textbf{0.384 $\pm$ 0.001} & \textbf{0.366 $\pm$ 0.001} \\
\bottomrule
\end{tabular}
\label{tab: mAP-small}
\end{table*}

\begin{table*}[ht]
\centering
\caption{Performance comparison for cross-scale transferability of synthetic data (small $\rightarrow$ medium/large).}

\begin{subtable}{\textwidth}
\centering
\caption{YOLOv11-medium performance.}
\begin{tabular}{lccc}
\toprule
Train dataset & Count & mAP & mAP w/o TP \\
\midrule
train-\textit{D} & 3,187 & 0.366 $\pm$ 0.004 & - \\
train-\textit{D} + naive v0 & 19,122 & 0.426 $\pm$ 0.002 & - \\
\hline
train-\textit{D} + naive v0 + naive v1 & 35,057 & \underline{0.429 $\pm$ 0.004} & \underline{0.415 $\pm$ 0.005} \\
train-\textit{D} + naive v0 + manual v1 & 35,057 & 0.428 $\pm$ 0.005 & 0.414 $\pm$ 0.005 \\
train-\textit{D} + naive v0 + automatic v1 & 35,057 & \textbf{0.429 $\pm$ 0.003} & \textbf{0.416 $\pm$ 0.005} \\
\bottomrule
\end{tabular}
\end{subtable}

\vspace{1em}

\begin{subtable}{\textwidth}
\centering
\caption{YOLOv11-large performance.}
\begin{tabular}{lccc}
\toprule
Train dataset & Count & mAP & mAP w/o TP \\
\midrule
train-\textit{D} & 3,187 & 0.371 $\pm$ 0.001 & - \\
train-\textit{D} + naive v0 & 19,122 & 0.419 $\pm$ 0.001 & - \\
\hline
train-\textit{D} + naive v0 + naive v1 & 35,057 & \underline{0.431 $\pm$ 0.003} & \underline{0.414 $\pm$ 0.004} \\
train-\textit{D} + naive v0 + manual v1 & 35,057 & 0.429 $\pm$ 0.007 & 0.411 $\pm$ 0.008 \\
train-\textit{D} + naive v0 + automatic v1 & 35,057 & \textbf{0.432 $\pm$ 0.005} & \textbf{0.415 $\pm$ 0.006} \\
\bottomrule
\end{tabular}
\end{subtable}
\label{tab: mAP-transfer}
\end{table*}

We aim to validate whether our proposed pipeline effectively generates edge-case data that addresses the blind spots of a given discriminative model. We primarily compare the efficacy of different prompt generation strategies for augmenting a training dataset via the T2I model. Due to space constraint, details on model architectures and training configurations are provided in Appendix~\ref{sec: model_spec}.

\subsection{Setup}

\paragraph{Comparison groups.} 
To evaluate the effectiveness of our proposed method as a prompt generation strategy, we compare it against the two baselines. For clarity, we present both the baselines and our method below:
\begin{itemize}
    \item \textbf{naive}: This baseline directly uses the base captions ($c_{base}$) generated by InternVL3-38B \citep{zhu2025internvl3} from the training images. It generates new images without any further rephrasing or modification to these captions, varying only the random seed of the T2I model for diversity.
    \item \textbf{manual}: Following prior work \citep{kim2025edge}, this baseline relies on manually-engineered prompts. These prompts are derived from detailed error analyses by AI researchers and are designed to elicit specific, known failure modes.
    \item \textbf{automatic}: Our method employs a preference-tuned LLM to automatically rephrase captions into descriptions of edge-case scenarios, removing the need for manual intervention.
\end{itemize}

\paragraph{Dataset.} We conduct our experiments on the FishEye8K dataset \citep{gochoo2023fisheye8k}, a benchmark adopted to the AI-City Challenge. The dataset consists of images from 20 fisheye cameras with annotations for five categories: Bus, Bike, Car, Pedestrian, and Truck. Given its inherent scene biases across diverse urban environments, FishEye8K serves as an optimal testbed for evaluating the effectiveness of our automated edge-case synthesis.

To rigorously assess model robustness, we partition the official training set into two disjoint subsets: train-\textit{D} and train-\textit{R}. To simulate a restricted training environment, train-\textit{D} is used for training the discriminative model while train-\textit{R} is used for the preference learning of the rephrasing LLM and contains a more diverse distribution. This setup allows the LLM to learn under-represented characteristics and generate corresponding synthetic data to fill the gaps in train-\textit{D}. Please refer to Appendix~\ref{sec: data_bias} for more details.

To prevent potential data leakage, the LLM is restricted from accessing train-\textit{D} during the preference learning phase and only interacts with it at inference time for data augmentation. This protocol ensures that the synthesized samples reflect the LLM's generalization capability rather than simple memorization. For completeness, the test set is held out exclusively for final evaluation.

\paragraph{Evaluation metrics.} We evaluate the effectiveness of the prompt generation strategies using the mAP gain of the discriminative model after augmenting the dataset with synthesized samples. This evaluation protocol allows us to directly quantify how much each augmentation strategy contributes to improving detection performance beyond the original training data. To explicitly assess how well each method addresses the model's blind spots, we introduce a custom metric, termed \textit{mAP w/o TP}, described in Appendix~\ref{sec: custom_metric}. In brief, this metric computes mAP only on ground-truth annotations that do not overlap with the True Positive predictions of the discriminative model trained on train-\textit{D} and naive v0, thereby focusing exclusively on previously misrecognized or challenging instances. A higher mAP w/o TP therefore indicates that the synthetic data more effectively improves the model's initial blind spots, rather than merely reinforcing patterns that the model has already learned.

\subsection{Visual diversity}

Figure~\ref{fig: syn_examples} provides a controlled comparison of how different prompt generation strategies influence synthetic visual outputs. By utilizing the same base caption for each row, we can directly observe the impact of rephrasing on synthesis.

While the automatic approach does not necessarily generate a higher number of object instances than the manual variant, it exhibits significantly greater background diversity, encompassing a wider range of lighting conditions, camera viewpoints, and scene contexts. This suggests that the preference-tuned LLM successfully explores a broader distribution of plausible scenarios rather than sticking to the main modes of the training distribution. In contrast, the manual baseline often produces repetitive scenes with limited background variation. Consequently, the automatic method is more effective at generating novel samples that cover underrepresented regions of the data distribution, thereby helping to alleviate dataset bias and improve model robustness

\subsection{Model performance}

As summarized in Table~\ref{tab: mAP-small}, our proposed method consistently outperforms both baselines across all experimental settings. To establish a competitive baseline, we define v0 as the initial iteration of data augmentation, where the train-\textit{D} is expanded five-fold using the naive strategy. This naive v0 dataset serves as the common starting point for all subsequent augmentation experiments. When augmenting the dataset further in the next iteration (v1), our method achieves the highest mAP and, more importantly, the highest mAP w/o TP.\footnote{Note that the model trained on train-\textit{D} + naive v0 serves as the reference for calculating the mAP w/o TP. Accordingly, we measure this metric exclusively for subsequent iterations of data augmentations.}

This result demonstrates that the data generated by our pipeline is more effective at correcting the baseline model's specific weaknesses compared to both naively generated data and data from manually engineered prompts. In contrast, other baselines show diminishing returns, suggesting that simply increasing data volume without targeting specific weaknesses is a less efficient strategy. For a class-wise and tag-wise analysis of the model performance to inspect the effect on dataset bias, please refer to Appendix~\ref{sec: class-wise} and \ref{sec: tag-wise}, respectively.

\subsection{Transferability}

To assess whether a dataset deemed effective for one model remains valid for another, a core concern in data-centric evaluation, we conducted transferability experiments. This approach aligns with recent findings \citep{ji2023randomness, go2023transferable} that emphasize robust evaluation designs, acknowledging that instance informativeness may vary across model configurations and necessitating experimental methods to identify broadly transferable data. To examine the transferability of synthetic data across model scales, we trained YOLOv11-medium and YOLOv11-large using datasets generated by the proposed pipeline with YOLOv11-small. As summarized in Table~\ref{tab: mAP-transfer}, the automatic still achieves the highest mAP and mAP w/o TP.

\begin{figure}[t]
    \centering
    \includegraphics[width=\linewidth]{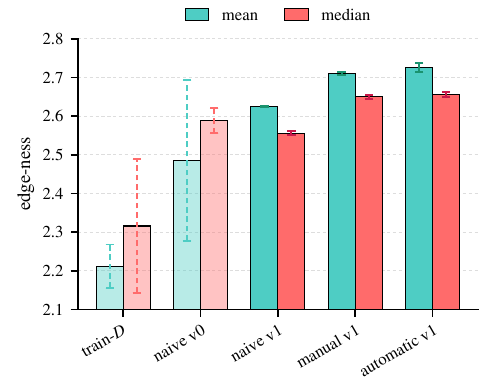}
    \caption{Comparison of edge-ness statistics across different data augmentation strategies. The proposed automatic pipeline produces samples with the highest mean and median loss, indicating more challenging training examples for the discriminative model.
    }
    \label{fig: edge-ness}
\end{figure}

\begin{figure*}[t]
    \centering
    \begin{subfigure}{0.23\textwidth}
        \centering
        \includegraphics[width=\linewidth]{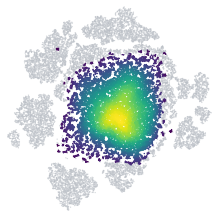}
        \caption{naive}
        \label{fig: naive}
    \end{subfigure}
    \hspace{0.06\textwidth}
    \begin{subfigure}{0.23\textwidth}
        \centering
        \includegraphics[width=\linewidth]{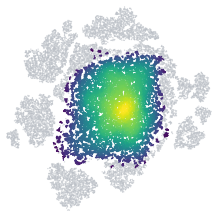}
        \caption{manual}
        \label{fig: manual}
    \end{subfigure}
    \hspace{0.06\textwidth}
    \begin{subfigure}{0.23\textwidth}
        \centering
        \includegraphics[width=\linewidth]{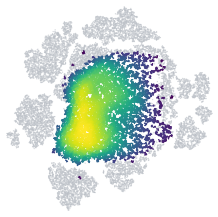}
        \caption{automatic}
        \label{fig: automatic}
    \end{subfigure}
    \caption{Comparison of naive, manual, and automatic by UMAP of CLIP embeddings. Gray points represent real data, while the colored ones indicate synthetic data, with colors closer to yellow denoting higher density.}
    \label{fig: embedding}
\end{figure*}

\subsection{Edge-ness}

We empirically validate the core mechanism of our pipeline: its ability to generate images with high edge-ness. We measure the training loss $\mathcal{L}_{task}$ of the discriminative model on data generated by each method. Following the standard training objective of our YOLOv11-small discriminator, $\mathcal{L}_{task}$ is specifically defined as the weighted sum of the bounding box regression loss, classification loss, and distribution focal loss. \citep{yolo11_ultralytics}. As shown in Figure~\ref{fig: edge-ness}, data from our proposed pipeline consistently induces the highest mean and median loss. This confirms that our preference learning framework successfully steers the LLM to generate captions that translate into verifiably more difficult examples for the discriminative model, thereby creating effective training signals for improving robustness.

\subsection{Embedding}

To better understand the underlying properties of the generated data, we visualize the image embeddings obtained by CLIP \citep{radford2021learning} for each generation strategy using UMAP \citep{mcinnes2018umap-software}, as shown in Figure~\ref{fig: embedding}. This visualization provides an intuitive view of how different strategies explore the feature space induced by a strong vision-language model. Compared to the naive approach, the manual strategy exhibits a slightly broader spread, indicating a modest increase in diversity. In contrast, the automatic method not only achieves a significantly greater spread, but also shifts its mode toward a sparser region of the real data embedding space. This shift suggests that the automatic strategy is more effective at generating novel and less frequently observed samples, potentially covering underrepresented regions of the data distribution. Such behavior is consistent with our objective of synthesizing edge-case examples that complement the existing training data.

\subsection{Linguistic analysis}

A closer inspection of the prompts and images of high-density regions reveals that our preference-tuned LLM learns a sophisticated rephrasing strategy. Compared to the factual source captions, its outputs tend to: (1) employ more descriptive, literary language; (2) emphasize actions and dynamic situations over static object descriptions; and (3) transform objective metadata (e.g., time, weather) into evocative descriptions of mood and atmosphere. This contrasts sharply with the manual baseline, where manual prompts are narrowly focused on object attributes. We interpret this as an automated strategy to enrich background context and scene diversity, which directly counteracts the inherent biases of the FishEye8K dataset. Please refer to Appendix~\ref{sec: pattern} for more structured analysis.

\section{Conclusion}
\label{sec:conclusion}

In this paper, we propose a novel pipeline for automated, text-guided edge-case synthesis.
This framework facilitates the creation of more diverse and challenging synthetic datasets, providing a scalable pathway to improve both training efficiency and generalization. We anticipate that the benefits of our pipeline will become even more pronounced as more advanced T2I models and pseudo-labelers emerge. Furthermore, considering dataset bias even during the fine-tuning of T2I models and pseudo-labelers could have improved performance. For future work, we plan to extend our study to a broader range of datasets and explore alternative uncertainty measures for edge-ness.
{
    \small
    \bibliographystyle{ieeenat_fullname}
    \bibliography{main}
}

\clearpage
\onecolumn
\appendix
\section{Data augmentation pipeline}
\label{sec: aug_pipeline}

\begin{figure}[h]
    \centering
    \includegraphics[width=\linewidth]{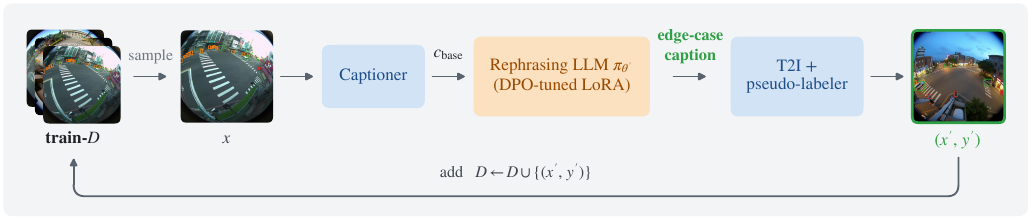}
    \caption{Data augmentation pipeline by inference of the preference-tuned LLM.}
    \label{fig: pipeline_aug}
\end{figure}

This diagram depicts the data augmentation process using the preference-tuned LLM obtained from the training phase. An image is sampled from the train-\textit{D} split and a base caption is generated. The preference-tuned LLM then produces the optimized "edge-case oriented" captions based on its learned preference. This targeted captions are used to synthesize the new, high-value images via the T2I model. After being assigned pseudo-annotations by a pseudo-labeler, this new synthetic data is added to the train-\textit{D} set, thereby augmenting the original training data with a sample specifically designed to address a model weakness.
\section{Model architectures and training configurations}
\label{sec: model_spec}

We summarize the details of the model architectures and training procedures employed in our paper.

\paragraph{Models.}  
We used five distinct models covering captioning, re-phrasing, generation, pseudo-labeling, and discriminative tasks:
\begin{itemize}
    \item Image captioning: InternVL3-38B \citep{zhu2025internvl3}
    \item Caption re-phrasing: Llama-3-8B-Instruct \citep{llama3modelcard}
    \item Text-to-image generation: Flux.1-dev \citep{flux2024}
    \item Pseudo-labeling: Co-DETR \citep{zong2023detrs}
    \item Discriminative: YOLOv11-small \citep{yolo11_ultralytics}
\end{itemize}
It is worth noting that the model selection closely aligns with the manual baseline \citep{kim2025edge}, except for the caption re-phrasing component. Unlike the baseline, which relied on the closed-source GPT-4.1-mini \citep{openai_gpt41_2024}, we employed an open-source model to allow fine-tuning for our task.

\paragraph{Training strategies.}  
The image captioning model (InternVL3-38B) was directly adopted from its publicly available pre-trained checkpoint in \texttt{transformers} \citep{Wolf_Transformers_State-of-the-Art_Natural_2020} without further fine-tuning. The caption re-phrasing model (Llama-3-8B-Instruct) was fine-tuned with the preference dataset constructed from train-\textit{R} using the \texttt{trl} \citep{von_Werra_TRL_Transformer_Reinforcement} with a Low-Rank Adaptation (LoRA) scheme. The text-to-image generation model (Flux.1-dev) was trained with train-\textit{D} via the \texttt{simpletuner} library, employing Low-Rank Kronecker product (LoKr) scheme. For the pseudo-labeling stage, we fine-tuned Co-DETR with train-\textit{D} using the \texttt{mmdetection} \citep{MMDetection_Contributors_OpenMMLab_Detection_Toolbox_2018} with full model finetuning. Finally, the discriminative model (YOLOv11-small) was trained with train-\textit{D} and synthetic datasets through full finetuning using the \texttt{ultralytics} \citep{Jocher_Ultralytics_YOLO_2023}.

\paragraph{Configurations.}  
All training was performed by the default configurations provided by each library, unless otherwise specified. We adopt most settings from prior work \citep{kim2025edge}, and release all configuration files, bash scripts, and step-by-step guides for reproducing the experiments at the \href{https://github.com/gokyeongryeol/ATES}{github repository}.

\paragraph{Hardware.}
All training and inference were performed on four NVIDIA A100 GPUs. The heaviest GPU usage occurred during image captioning with InternVL3-38B, which required approximately 75GiB VRAM per device.
\section{Analysis of bias in data splits}
\label{sec: data_bias}

\begin{figure*}[t]
    \centering
    \includegraphics[width=0.8\linewidth]{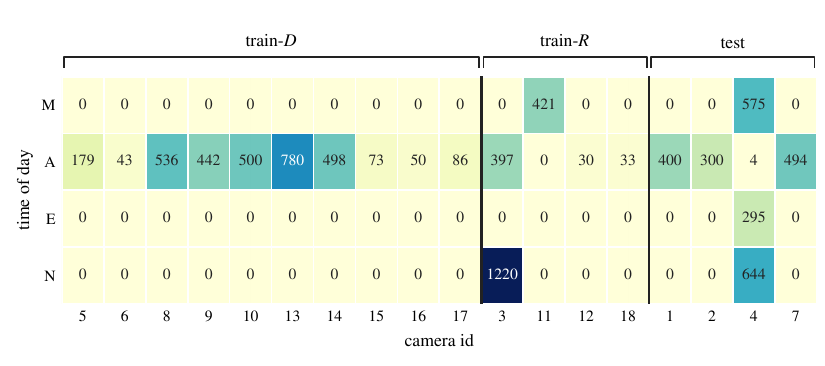}
    \caption{Number of images per camera ID and time-of-day in FishEye8K. The dataset is split into train-\textit{D}, train-\textit{R}, and test sets based on camera IDs, intentionally creating different levels of bias across splits.}
    \label{fig: data_dist}
\end{figure*}

We divided the FishEye8K dataset into three subsets based on camera IDs to intentionally introduce dataset biases and to study their impact on discriminative and generative modeling. The splits are as follows:

\begin{itemize}
    \item train-\textit{D}: camera ids \{5, 6, 8, 9, 10, 13, 14, 15, 16, 17\}
    \item train-\textit{R}: camera ids \{3, 11, 12, 18\}
    \item test: camera ids \{1, 2, 4, 7\}
\end{itemize}

As shown in Figure~\ref{fig: data_dist}, the distribution of images across time-of-day and camera IDs reveals distinct patterns of bias. Specifically, the train-\textit{D} split contains images exclusively from the afternoon (A), with no images from morning (M), evening (E), or night (N). In contrast, the train-\textit{R} split exhibits a moderate bias with images from multiple times of day, while the test split is relatively unbiased.

This deliberate separation allows us to train a discriminative model on the highly biased train-\textit{D} split, highlighting its limitations when generalizing to other times of day. Meanwhile, the train-\textit{R} split serves to allow the rephrasing LLM to automatically capture under-represented characteristics of train-\textit{D} split and generate such data.
\section{Custom evaluation metric proposal}
\label{sec: custom_metric}

\subsection{Background and motivation}
In data-centric AI, a critical task is to identify a model's weaknesses and subsequently verify whether new algorithms or data effectively address them. In object detection, these weaknesses often manifest as instance-level failures, such as objects that are occluded, truncated, or small. These \textit{weak instances} are typically rare compared to easily detectable ones. Consequently, global performance metrics like mean Average Precision (mAP) are often insensitive to improvements on these specific blind spots, as the metric's overall fluctuation is dominated by the vast number of \textit{easy} instances. This makes it challenging to determine if a new approach genuinely targets and resolves the model's key deficiencies.

\subsection{Limitations of existing analysis methods}
Several methods are commonly used to identify model weaknesses, but each has its limitations:
\paragraph{Class-wise AP} While useful for identifying underperforming classes, it is not sufficiently informative when a class has large intra-class variation and the model consistently fails on specific patterns within that class.

\paragraph{False positive/negative analysis} Visualizing the correction of false positives and false negatives provides qualitative evidence of improvement. However, there is no standard quantitative metric to measure this change systematically across the entire dataset.

\paragraph{Embedding and uncertainty analysis} Techniques that identify sparse regions in an embedding space or high-uncertainty samples are often used as proxies for data difficulty. However, these methods may require separate, large-scale models \citep{radford2021learning, oquab2023dinov2}. While effective in general domains, these models often require additional fine-tuning for specialized domains (e.g., manufacturing defects, medical imaging), limiting their direct applicability.

\subsection{Proposal: mAP w/o TP}
To address these limitations, we propose \textbf{mAP w/o TP} (mean Average Precision without True Positives of a baseline model). This metric provides a clear, quantitative way to measure how effectively a new model (Model $\mathcal{B}$) improves upon the specific weaknesses identified by a baseline model (Model $\mathcal{A}$).

\begin{figure}[ht]
    \centering
    \includegraphics[width=0.95\linewidth]{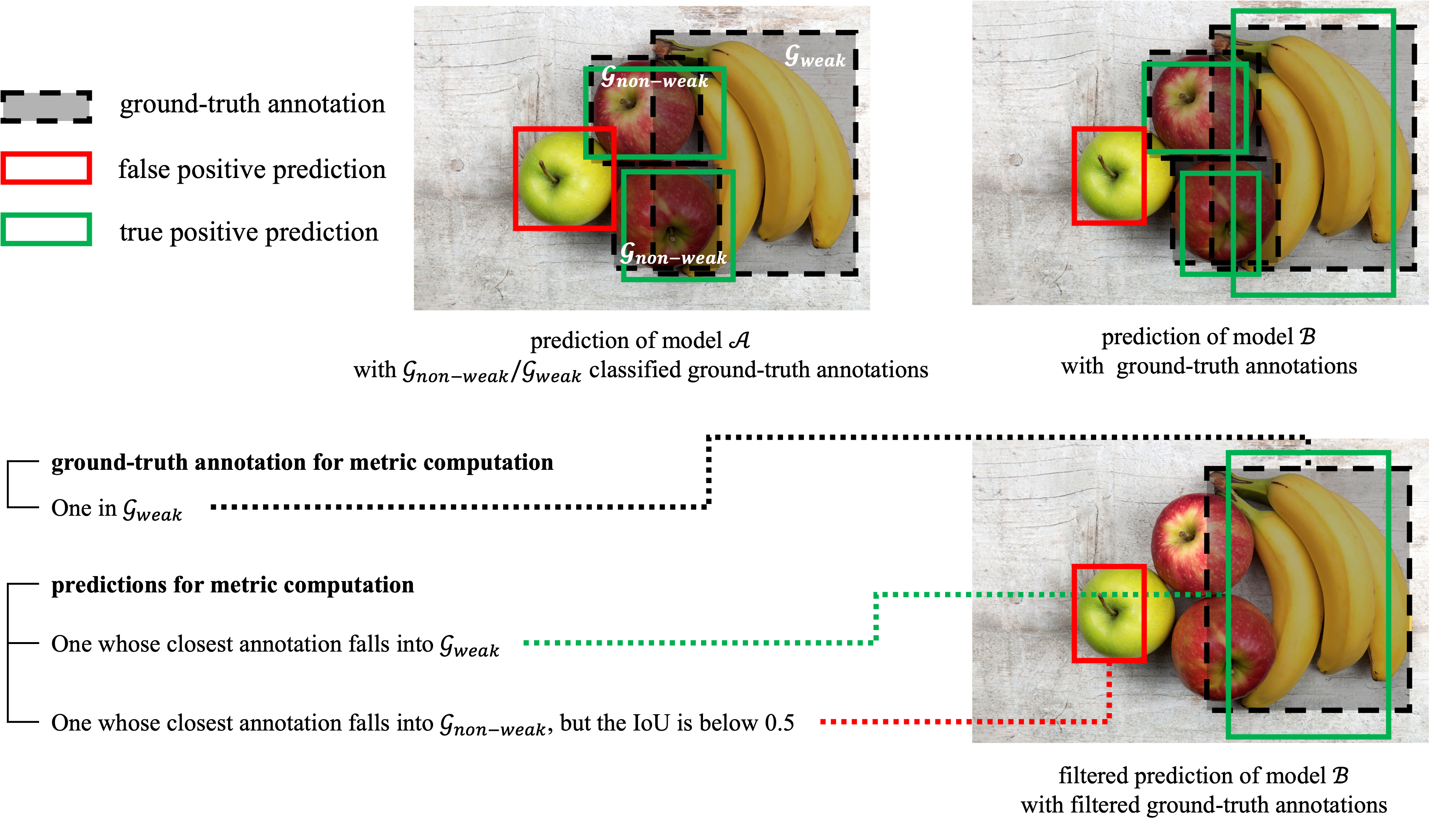}
    \caption{Filtered ground-truth annotations and predictions to compute mAP w/o TP.}
    \label{fig: map_wo_TP}
\end{figure}

The core idea is to first define \textit{non-weak instances} as the set of annotations that the baseline Model $\mathcal{A}$ already detects correctly (its true positives (TP)). Then, when evaluating Model $\mathcal{B}$, we exclude its correct predictions on these non-weak instances from the calculation. This focuses the evaluation squarely on the subset of data that was challenging for the baseline model, making any improvements on these blind spots much more apparent. We provide an example of the filtered ground-truth annotations and predictions in Figure~\ref{fig: map_wo_TP}.

\subsection{Step-by-Step calculation}
The mAP w/o TP for a new Model $\mathcal{B}$ relative to a baseline Model $\mathcal{A}$ is calculated as follows:
\begin{enumerate}
    \item Inference with baseline model: First, run inference with Model $\mathcal{A}$ on the evaluation dataset to obtain its set of predictions.

    \item Partition ground-truth annotations: Categorize all ground-truth annotations in the evaluation set into two disjoint sets based on Model $\mathcal{A}$'s predictions:
    \begin{itemize}
        \item $\mathcal{G}_{\text{non-weak}}$: The set of annotations correctly identified by a TP prediction from Model $\mathcal{A}$ (based on the IoU greater than 0.95 and class label matching).
        \item $\mathcal{G}_{\text{weak}}$: Other annotations, which correspond to Model $\mathcal{A}$'s FN or were not associated with any high-confidence prediction.
    \end{itemize}

    \item Inference with new model: Run inference with the new Model $\mathcal{B}$ on the same evaluation dataset to obtain its predictions.

    \item Filter predictions of new model: Filter the predictions from Model $\mathcal{B}$ based on their relationship with the partitioned annotation sets. A prediction from Model $\mathcal{B}$ is kept for evaluation only if it meets one of the following criteria:
    \begin{itemize}
        \item Its highest-IoU corresponding annotation belongs to $\mathcal{G}_{\text{weak}}$ (i.e., it addresses a weak instance).
        \item Its highest-IoU corresponding annotation belongs to $\mathcal{G}_{\text{non-weak}}$, but the IoU is less than 0.5. This penalizes cases where Model $\mathcal{B}$ fails on an easy instance that Model $\mathcal{A}$ handled correctly.
    \end{itemize}
    \item Calculate final metric: Calculate the mAP, class-wise AP, or other relevant metrics using only the filtered set of predictions from the previous step and ground-truth annotations in $\mathcal{G}_{\text{weak}}$.
\end{enumerate}

\subsection{Significance and benefits}
The mAP w/o TP metric offers several key advantages:
\begin{itemize}
    \item It provides a \textbf{quantitative and direct} measure of a model's ability to resolve the specific blind spots of a baseline.
    \item By isolating the evaluation to weak instances, it makes the impact of targeted algorithms or data augmentation strategies \textbf{more interpretable}.
    \item It is \textbf{model-agnostic} and universally applicable. It does not depend on a model's architecture or training methodology and can be used to compare any two models.
\end{itemize}
\section{Analysis in a class-wise manner}
\label{sec: class-wise}

\begin{table}[ht]
    \centering
    \caption{Per-class statistics and evaluation performance.}
    \vspace{0.3em}

    \begin{subtable}[h]{\linewidth}
        \centering
        \begin{tabular}{lccccc}
            \toprule
            Setting & Bus & Bike & Car & Pedestrian & Truck \\
            \midrule
            train-D + naive v0 + naive v1 & 8,904 & \underline{247,519} & \textbf{205,875} & 60,201 & 4,064 \\
            train-D + naive v0 + manual v1 & \textbf{11,916} & 232,935 & 202,669 & \underline{69,967} & \textbf{7,367} \\
            train-D + naive v0 + automatic v1 & \underline{9,457} & \textbf{250,853} & \underline{203,388} & \textbf{86,666} & \underline{4,230} \\
            \bottomrule
        \end{tabular}
        \caption{Per-class instance counts}
    \end{subtable}

    \vspace{1em}

    \begin{subtable}[ht]{\linewidth}
        \centering
        \begin{tabular}{lccccc}
            \toprule
            Setting & Bus & Bike & Car & Pedestrian & Truck \\
            \midrule
            train-D + naive v0 + naive v1 & 0.509 & 0.303 & \textbf{0.507} & 0.189 & \textbf{0.368} \\
            train-D + naive v0 + manual v1 & \textbf{0.527} & 0.308 & 0.503 & \underline{0.218} & 0.332 \\
            train-D + naive v0 + automatic v1 & \underline{0.510} & \textbf{0.320} & \underline{0.505} & \textbf{0.221} & \underline{0.343} \\
            \bottomrule
        \end{tabular}
        \caption{Per-class evaluation performance (mAP)}
    \end{subtable}
    \label{tab: class-wise}
\end{table}

We report the class-wise instance counts and detection performance of YOLOv11-small \citep{Jocher_Ultralytics_YOLO_2023} in Table~\ref{tab: class-wise}. The results indicate that the automatic provides more balanced improvements across classes compared to the naive and manual baselines. In particular, the Pedestrian class, originally underrepresented and difficult to detect, shows the largest gap. Similarly, the Bike class, which had a large number of training samples but relatively low accuracy, also benefits from the automatic strategy. In short, the automatic effectively mitigates this trade-off by improving rare and difficult classes (Pedestrian, Bike) while maintaining stable performance on other categories.

In contrast, the naive primarily increases data quantity, which boosts performance for dominant classes such as Car, but fails to consistently improve underrepresented categories like Pedestrian and Bike. This suggests that quantity alone does not guarantee higher performance when the additional data lacks diversity and edge-ness. While the manual improves Pedestrian accuracy, it comes at a cost, markedly reducing the Truck score despite a substantial increase in its instance count.

\section{Analysis in a tag-wise manner}
\label{sec: tag-wise}

To assess the impact on dataset bias, we conduct a tag-wise analysis with respect to the time of day: morning, evening, and night. These categories are particularly critical as they represent an underrepresented tag in the dataset. As shown in Table~\ref{tab: tag-wise}, although naive and manual strategies yield gains in certain cases, the automatic approach consistently achieves equal or superior performance across the time of days. It is worth noting that the manual baseline serves as a strong baseline in this evaluation, given its design for fine-grained error analysis, even though it often underperforms the naive baseline in other quantitative results. In contrast, the automatic not only adapts more effectively to rare conditions but also delivers more robust improvements overall. These findings highlight the automatic strategy as a more effective solution for alleviating dataset bias in underrepresented tags.

\begin{table}[ht]
\centering
\caption{Performance comparison across time of day.}
\begin{tabular}{lccc}
\toprule
Setting & M (morning) & E (evening) & N (night) \\
\midrule
train-\textit{D} + naive v0 + naive v1 & \textbf{0.402 $\pm$ 0.002} & 0.497 $\pm$ 0.016 & 0.177 $\pm$ 0.013 \\
train-\textit{D} + naive v0 + manual v1 & 0.387 $\pm$ 0.014 & \underline{0.502 $\pm$ 0.011} & \underline{0.182 $\pm$ 0.020} \\
train-\textit{D} + naive v0 + automatic v1 & \underline{0.398 $\pm$ 0.009} & \textbf{0.509 $\pm$ 0.007} & \textbf{0.183 $\pm$ 0.007} \\
\bottomrule
\end{tabular}
\label{tab: tag-wise}
\end{table}

\section{Analysis of rephrasing patterns}
\label{sec: pattern}

We analyze 20 caption--rephrased pairs located in high-density regions of Figure~\ref{fig: automatic}. This analysis reveals several consistent patterns that highlight the strategies used to enrich the original descriptions and make them more vivid and informative. The six most prominent trends are as follows:

\begin{enumerate}
    \item \textit{Scene atmosphere enhancement}: Rephrased captions frequently introduce emotive and descriptive adjectives, such as ``serene'', ``peaceful'', ``bustling'', or ``vibrant'', to convey the mood of the scene more vividly than the original caption. This trend appears in all 20 pairs, indicating its universal role in enhancing expressiveness.

    \item \textit{Perspective specification}: Many rephrasings explicitly indicate the camera viewpoint (e.g., ``front-view'', ``side-view'', ``low-angle''), adding spatial context that is often implicit in the original captions. This occurs in the majority of pairs (15/20), highlighting the importance of visual perspective in rephrasing.

    \item \textit{Temporal and weather visualization}: Neutral references to weather or time (e.g., ``clear sky'', ``daytime'') are often expanded into more expressive phrases, such as ``golden morning light'', ``sun-drenched afternoon'', or ``warm glow'', enhancing the visual imagery. This trend is consistently applied across all pairs, showing a clear preference for temporal and environmental enrichment.

    \item \textit{Intersection type clarification}: Generic mentions of ``intersection'' are frequently replaced with more precise terms like ``T-junction'', ``Y-junction'', or ``street corner'', providing clearer spatial context. Applied in 14/20 pairs, this trend improves spatial specificity without altering the core scene description.

    \item \textit{Action emphasis}: Rephrased captions tend to highlight the dynamics of pedestrians and vehicles, converting simple existence statements into descriptive actions such as ``stroll'', ``zip by'', or ``weave through'', thereby increasing narrative engagement. This occurs in 16/20 pairs, emphasizing the narrative benefit of describing movement.

    \item \textit{Urban context enrichment}: Additional details about shops, cafes, buildings, and signage are often added to create a richer urban setting, making the scene more specific and relatable. This trend is applied in 17/20 pairs, reflecting a strong tendency to augment environmental details.
\end{enumerate}

Overall, these trends consistently appear across the majority of caption pairs, suggesting that effective rephrasing not only preserves the factual content but also enhances \textit{expressiveness}, \textit{spatial awareness}, and \textit{narrative dynamics}.

\section{Prompt templates}
\label{sec: prompt}

We present the exact prompts used in our experiments. In particular, we list them in sequence: starting with the prompt used by InternVL3-38B to generate captions from real images, followed by those applied with Llama-3-8B-Instruct for caption rephrasing under the manual baseline, for constructing the preference dataset, and with the preference-tuned LLM.

\subsection{Image captioning prompt}

\begin{tcolorbox}[
    colback=gray!5,
    colframe=gray!50,
    boxrule=0.8pt,
    arc=4pt,
    left=6pt,
    right=6pt,
    top=6pt,
    bottom=6pt,
    enhanced,
    breakable,
]
You are an expert in generating high-quality image captions. Please analyze the provided fish-eye image in detail.\\[0.8em]
Please adhere to the following format for the caption:
\begin{itemize}[leftmargin=1.5em]
    \item Start with ``A photo of''.
    \item Limit the total length to 40-50 words.
    \item  Focus on bus, bike, car, pedestrian and truck.
    \item  Describe time of day, weather and location.
    \item  Focus on the scene inside the fish-eye lens.
    \item  Use grammatically correct and clear sentences.
\end{itemize}
\end{tcolorbox}

\subsection{Rephrasing prompt by the manual baseline}
\label{sec:prompt-manual}
\begin{tcolorbox}[
    colback=gray!5,
    colframe=gray!50,
    boxrule=0.8pt,
    arc=4pt,
    left=6pt,
    right=6pt,
    top=6pt,
    bottom=6pt,
    enhanced,
    breakable,
]
You are an expert at visually-grounded image caption rewriting. Your task is to rewrite image captions taken with a fisheye camera so that:
\begin{itemize}[leftmargin=1.5em]
    \item The objects Bus, Bike, Car, Pedestrian, and Truck are small in scale and located near the edges of the image, where fisheye distortion is strong.
    \item Object placement and interaction should be plausible within real-world traffic scenes.
\end{itemize}

Please adhere to the following format for the caption:
\begin{itemize}[leftmargin=1.5em]
    \item Start with ``A photo of''.
    \item Limit the total length to 40-50 words.
    \item Use grammatically correct and clear sentences.
\end{itemize}

Apply the following object descriptions:
\begin{itemize}[leftmargin=1.5em]
    \item Bus: large public passenger vehicles
    \item Truck: heavy-duty vehicles like dump trucks or semi-trailers
    \item Car: compact vehicles such as sedans, SUVs, or vans
    \item Bike: bicycles, motorcycles, or scooters, either parked or with riders
    \item Pedestrian: visible people walking, standing, or crossing
\end{itemize}

Avoid visual ambiguity between:
\begin{itemize}[leftmargin=1.5em]
    \item Bus vs Truck $\rightarrow$ contrast size, function, silhouette
    \item Car vs Truck $\rightarrow$ emphasize bulk and height differences
    \item Pedestrian vs Bike $\rightarrow$ distinguish by motion, vehicle presence, posture
\end{itemize}

Ensure variation across scene conditions:
\begin{itemize}[leftmargin=1.5em]
    \item Camera angles: side-view or front-view
    \item Intersection types: T-junctions, Y-junctions, cross-intersections, mid-blocks, pedestrian crossings, or irregular layouts
    \item Lighting: morning, afternoon, evening, or night
    \item Traffic flow: free-flowing, steady, or busy
\end{itemize}

When choosing scene elements, slightly favor the following (but still maintain diversity overall):
\begin{itemize}[leftmargin=1.5em]
    \item Categories prominently shown: Pedestrian and Truck
    \item Time of day: Day, Afternoon, Night, or general Daytime
    \item Weather: Clear
    \item Location type: Urban areas such as City streets
\end{itemize}

Preserve the core content of the original caption, but rewrite it to reflect the above constraints. If none of the specified categories are present, you may subtly introduce one or more at the distorted outer edges of the image. Always maintain natural, fluent language, and don't make the added objects the main focus unless already emphasized.
\end{tcolorbox}

\subsubsection{Rephrasing prompt to construct preference dataset}

\begin{tcolorbox}[
    colback=gray!5,
    colframe=gray!50,
    boxrule=0.8pt,
    arc=4pt,
    left=6pt,
    right=6pt,
    top=6pt,
    bottom=6pt,
    enhanced,
    breakable,
]
You are an expert in generating diverse yet realistic image captions for road scenes captured by a fish-eye surveillance camera. Your task is to rewrite the caption I give you into 5 diverse and realistic variants.\\[0.8em]
Please adhere to the following format for the caption:
\begin{itemize}[leftmargin=1.5em]
    \item Start with ``A photo of''.
    \item Limit the total length to 40--50 words.
    \item Use grammatically correct and clear sentences.
    \item While preserving the core elements (bus, bike, car, pedestrian, truck) of the original caption, vary:
    \begin{itemize}
        \item Camera angles: side-view or front-view
        \item Intersection types: T-junctions, Y-junctions, cross-intersections, mid-blocks, or pedestrian crossings
        \item Lighting: morning, afternoon, evening, or night
        \item Traffic flow: free-flowing, steady, or busy
        \item Scene content: object count/placement and background features (e.g., buildings, shops, trees, signs, utility poles)
    \end{itemize}
    \item Ensure each caption includes at least one distinct variation that differentiates it from the others.
    \item Output the 5 captions as a numbered list, using the format:
    \begin{itemize}
        \item Caption 1
        \item Caption 2
        \item Caption 3
        \item Caption 4
        \item Caption 5
    \end{itemize}
    \item Do not include any explanation or extra text outside of the list.
\end{itemize}
\end{tcolorbox}

\subsubsection{Rephrasing prompt with the preference-tuned LLM}

\begin{tcolorbox}[
    colback=gray!5,
    colframe=gray!50,
    boxrule=0.8pt,
    arc=4pt,
    left=6pt,
    right=6pt,
    top=6pt,
    bottom=6pt,
    enhanced,
    breakable,
]
You are an expert in generating diverse yet realistic image captions for road scenes captured by a fish-eye surveillance camera.
Your task is to rewrite the caption I give you into a realistic variant.\\[0.8em]
Please adhere to the following format for the caption:
\begin{itemize}[leftmargin=1.5em]
    \item Start with ``A photo of''.
    \item Limit the total length to 40-50 words.
    \item Use grammatically correct and clear sentences.
    \item While preserving the core elements (bus, bike, car, pedestrian, truck) of the original caption, vary:
    \begin{itemize}
        \item Camera angles: side-view or front-view
        \item Intersection types: T-junctions, Y-junctions, cross-intersections, mid-blocks, or pedestrian crossings
        \item Lighting: morning, afternoon, evening, or night
        \item Traffic flow: free-flowing, steady, or busy
        \item Scene content: object count/placement and background features (e.g., buildings, shops, trees, signs, utility poles)
    \end{itemize}
\end{itemize}
\end{tcolorbox}

\end{document}